%% file: main.tex
\documentclass[conference]{IEEEtran}
\IEEEoverridecommandlockouts

\usepackage{amsmath,amssymb,amsfonts}
\usepackage{algorithmic}
\usepackage{graphicx}
\usepackage{textcomp}
\usepackage{xcolor}
\def\BibTeX{{\rm B\kern-.05em{\sc i\kern-.025em b}\kern-.08em
    T\kern-.1667em\lower.7ex\hbox{E}\kern-.125emX}}
    
\usepackage{times}  %
\usepackage{helvet}  %
\usepackage{courier}  %
\usepackage[hyphens]{url}  %
\usepackage{caption} %
    
\graphicspath{{figures/}} %
    
\title{Extracting Explainable Dates From Medical Images By Reverse-Engineering UNIX Timestamps} 

\author{
    \IEEEauthorblockN{Lee Harris}
    \IEEEauthorblockA{
        The University of Kent, Canterbury, UK\\
        \& \\
        TMLEP Research, Ashford, UK\\
        lah46@kent.ac.uk
    }
}

\begin{document}
	\maketitle

    \begin{abstract}
		\input{sections/abstract}

    \end{abstract}


    \section{Introduction}
    \label{sec:introduction}
    \input{sections/introduction}

	\section{Background And Related Work}
    \label{sec:background}
    \input{sections/background}

    \section{Initial Attempts To Recognise Dates}
    \label{sec:initial_attempts_to_recognise_dates}

\input{sections/learning_limitations}

    \section{Recognising Dates Using Regular Expressions} 
    \label{sec:capturing_dates_from_machine_written_text}

\input{sections/capturing_dates_from_machine_written_text}

    \section{Acknowledgements}
    \label{sec:acknowledgements}
    \input{sections/acknowledgements}

    \section{Conclusion}
    \label{sec:conclusion}

\input{sections/conclusion}

    \bibliography{IEEEabrv,ktp}
\end{document}

%% file: sections/abstract.tex
Dates often contribute towards highly impactful medical decisions, but it is rarely clear how to extract this data.
AI has only just begun to be used transcribe such documents, and common methods are either to trust that the output produced by a complex AI model, or to parse the text using regular expressions.
Recent work has established that regular expressions are an explainable form of logic, but it is difficult to decompose these into the component parts that are required to construct precise UNIX timestamps. %
First, we test publicly-available regular expressions, and we found that these were unable to capture a significant number of our dates. 
Next, we manually created easily-decomposable regular expressions, and we found that these were able to detect the majority of \textit{real} dates, but also a lot of sequences of text that \textit{looked} like dates. 
Finally, we used regular expression synthesis to automatically identify regular expressions from the reverse-engineered UNIX timestamps that we created. 
We find that regular expressions created by regular expression synthesis detect far fewer sequences of text that look like dates than those that were manually created, at the cost of a slight increase to the number of missed dates. 
Overall, our results show that regular expressions can be created through regular expression synthesis to identify complex dates and date ranges in text transcriptions. %
To our knowledge, our proposed way of learning deterministic logic by reverse-engineering several many-one mappings and feeding these into a regular expression synthesiser is a new approach. %

%% file: sections/introduction.tex
A very typical way to extract data from medical records (which continue to primarily be stored as paper documents) is to use human labour to manually transcribe these into digital representations, and to identify key points of interest (such as dates, names, medications, outcomes, etc.) at the same time.
However, this is a costly and error-prone process, and identifying employees who can (and will) undertake this mundane work may be difficult \cite{marx2009long}.
This may also need to be repeated if the required data to extract were to change.
It is for these reasons that being able to automatically transcribe and capture data would be very beneficial.

Dates are one type of data that is incredibly useful to obtain.
\cite{mandal2015date} and \cite{zhenwei2016datefinder} achieved this by using modern computer vision. 
However, as we highlight in Sec.~\ref{sec:initial_attempts_to_recognise_dates}, it is not be possible to identify the precise days, months, or years for complex dates or date ranges in this way due to the huge (potentially infinite) number of possible variations. 
Furthermore, it is rarely possible to explain (Sec.~\ref{ssec:explainability}) how these models reach their decisions, and this is an especially important consideration in high-stake situations (such as when handling medical data).
Information concluded from high-stake data is often highly impactful, and incorrect results are often  intensely scrutinised. 
The inability to explain how a decision were reached is one of the reasons why many governments require medical data to be processed by humans rather than AI \cite{veale2021demystifying}.

One way to achieve explainability is to use deterministic predicate rules of the form ``if [$\dots$] then [$\ldots$]’’.  
Other researchers and software packages, such as \cite{grolemund2011dates} and \footnote{\url{https://github.com/akoumjian/datefinder}}, achieve this by identifying date-like sequence of text using regular expressions (Sec.~\ref{ssec:regular_expressions}).
These are fast, but their use is often programming-language dependant. %
Additionally, dayless or multi-line dates, and words (e.g., `of', `to the', etc) that indicate date ranges, are rarely considered in other works.

We used regular expressions to capture and convert complex dates and date ranges present in real medical documents into precise UNIX timestamps. 
We achieved this by transcribing the documents using an open source Optical Character Recognition (OCR) and Handwritten Character Recognition (HCR) tool, and then pattern-matching against these using regular expressions.
Our first research contribution highlights that computer vision and Large Language Models (LLMs) are not able to solve this task. 
Our second contribution shows that easily-accessible regular expression patterns are not suitable for detecting dates and date ranges. 
Our third contribution shows that the number of correctly identified dates can be significantly increased by manually constructing regular expressions.
Finally, we show that all UNIX timestamps between two points in time can be captured by creating synthetic text containing these dates, and feeding the respective day, month, and year components into a regular expression synthesiser.
Using a regular expression synthesiser accounts for different programming languages, produces decomposable regular expressions, and ensures that constructing regular expressions is robust and consistent.

No medical data has been made available due to the sensitive nature of this, and so its characteristics are described instead. %

%% file: sections/background.tex
\subsection{Dates And UNIX Timestamps}
\label{ssec:unix_timestamps}
A date identifies a particular moment in time, in a specific timezone (often UTC or GMT), and this is often represented as the  amount of seconds that have elapsed since the 1st of January, 1970 (known as the epoch). 
These (UNIX, epoch, or POSIX timestamps) have become the standard way to digitally reason about, transfer, and store dates and times, and functionality for calculating these is ubiquitous. %
\cite{louis2020time} highlights that the Universal Coordinated Time (UTC) standard did not exist until 1970, so calculating negative UNIX timestamps is not an official part of the standard, but it is a common convention for negative numbers to be used when earlier dates are required.
Additionally, the technical specification used an unsigned 32-bit number, and as each second adds a value of 1, the integer overflow would limit the dates that can be represented to those at the epoch, and the end of 2038. 
This 69 year window is too short for many usecases, and so it is common for tools to use a 64 (or more) bit integer or floating-point representation.

\subsection{Optical and Handwritten Character Recognition}
\label{ssec:ocr_and_hcr}
Digital document transcriptions are created by visually recognising characters, and this is a historically significant research area \cite{mori1992historical}.
The text could exist in a range of file formats, such as in scanned images, photographs, electronically written documents, and mixed medium files (e.g., PDFs), %
and there exists an important distinction between Optical and Handwritten Character Recognition (respectively known as OCR and HCR).
Many researchers regard OCR as a ``solved'' problem  that can be achieved using far simpler software and hardware than HCR \cite{memon2020handwritten}.
Ch.~4 in~\cite{kaicheva2015process} lists many visual components of text that are more consistent in machine-written text than handwritten text, including: typeface, weight, size, skew, rotation, spacing, padding, sharpness, and colour. %
Various open-source and proprietary deep learning architectures have achieved incredible results in this field recently \cite{li2023trocr}, but it is rarely clear how  \cite{achtibat2024attnlrp}.

\subsection{Explainability}
\label{ssec:explainability}
Explainability is often seen as an essential requirement for responsible and ethical AI deployment in high-stake applications.
The Defense Advanced Research Projects Agency (DARPA) define this as ``AI systems that can explain their rationale to a human user, characterize their strengths and weaknesses, and convey an understanding of how they will behave in the future'' \cite{gunning2019darpa}.
Businesses in Europe are required to comply with the European Union's AI Act \cite{veale2021demystifying}, explainability is often required to debug and improve an AI model \cite{muggleton2018ultra}, and it is necessary to enforce accountability \cite{busuioc2021accountable}.
An AI model is typically seen as explainable if its inherent structure provides understandable justifications for its decisions or predictions regarding a particular data example, and rules of the form $`if \ldots then \dots'$ are often included in this definition \cite{schoenborn2021explainable}.
It is commonly accepted that inherently explainable AI models achieve noticeably poorer predictive accuracy than modern (i.e., deep-learning-based) algorithms though, and using intrinsically explainable algorithms instead of these 
is an active area of research \cite{roy2022tutorial}.
Many deep learning researchers suggest that explainability is the ability to extract functionally understandable representations (e.g., such as heatmaps when exploring images; \cite{montavon2018methods}) of the input data that humans will then understand, whilst others argue that inherently explainable AI models should be used from the offset \cite{rudin2019stop}.

\subsection{Regular Expressions}
\label{ssec:regular_expressions}
\begin{figure}[th]

\includegraphics[width=0.75\linewidth]{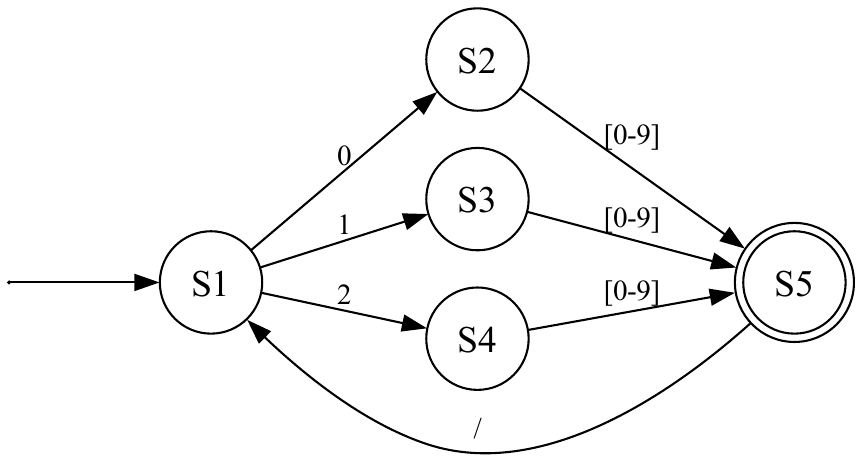}

\caption{A Deterministic Finite Automata (DFA) which would match a sequence of digit pairs separated by forward slashes. 
The digit pairs would consist of either a 0, 1, or 2, followed by any digit between 0 and 9 (i.e., [0-9]).
The start state is indicated by the arrow on the left side of the image (i.e., into S1), and the end state is shown by the circle nested inside of another circle (i.e., S5).
The equivalent regular expression to the DFA could be written as `$((0|1|2)[0-9]/)$*' or `$((0|1|2)\backslash$d\{1\}/)*' (where $\backslash$d\{1\} represents a positive digit containing a single character, and `*' represents any number of occurrences of the preceding pattern).
}

\label{fig:naive_dfa}
\end{figure}

Regular expressions (i.e., regexes) deterministically manipulate a body of text using character patterns. %
These may contain literal (e.g., `a', `b', `c') and/or meta (such as any base 10 digit) characters, and it is possible to detect lower or upper case text. 
Regexes may appear complex, but concise variants of these are often seen as an explainable form of AI \cite{roy2023learning} as they can be manually applied (albeit slowly) to a sequence of characters by humans.
It is also possible to covert these to clearer state machine visualisations. 
Fig.~\ref{fig:naive_dfa} shows a Deterministic Finite Automaton (DFA) state machine that would capture a (potentially) infinite sequence of digit pairs separated by forward slashes.

\subsection{Regular Expression Synthesis}
\label{ssec:existing_approaches}
Regular expression synthesis is a technique for automatically generating suitably compact and generalisable regular expressions. %
Some researchers have approached this from the perspective of loss minimisation, for instance the FOREST \cite{ferreira2021forest} algorithm is able to perform well in small n, large-p tasks by using an ensemble of decision trees, while genetic programming \cite{bartoli2016inference} has been shown to work particularly well when the input space is large.
However, these solutions are associated with typical learning problems (e.g., an inability to converge, computational cost, loss metric suitability, etc.), and identifying what should not be recognised is often incredibly difficult.
\cite{roy2023learning} highlight that an example of how not to drive a car would be the undesirable example of hitting a pedestrian. 
Alternatively, neural synthesisers \cite{chen2023data} can accept examples or natural language descriptions (e.g., ``create a regex that would [...]''), and the ubiquity of modern deep learning libraries and frameworks has produced models that can manipulate large amounts of data.
As reported by \cite{pertseva2022regex} though, these tend to create overly-generic or overly-specific regexes.
The approach used by \cite{avellaneda2019inferring} is similar to ours in that they created regular expressions by generating example data, but their goal was to produce a regular expression that correctly generalised to the entirety of a given language, whereas we want to restrict the regexes generalisation, as we can create most of the input space.

\subsection{Precision and Recall} %
Recall tells us how likely we are to miss the correct dates, while precision tells us how likely we are to predict dates that are not really there. 
Given that the true positives are the number of target dates that were predicted (by the regexes) as dates, the false positives are the number of target dates that were wrongly predicted, and the false negatives are the number of target dates that were not predicted, the precision and recall can formally be written as 

$$
	Precision = \frac{True Positives}{True Positives + False Negatives},
$$

\noindent{} and, 

$$
	Recall = \frac{True Positives}{True Positives + False Positives}
$$

These would usually incorporate the number of true negatives (i.e. all other document text), but it is not obvious how to quantify this. %

%% file: sections/learning_limitations.tex
\cite{zhenwei2016datefinder} successfully identified the positions of handwritten dates in a sequence of images, but they needed to enforce very rigid constraints (colour, style, position, year bound, size, language, date format, etc.) in order to do so, and knowing the position did not simultaneously produce the date parts (i.e., day, month, year) or UNIX timestamp.
Our initial investigations suggested that recognising dates using computer vision was not going to satisfy our requirements due to the huge (potentially infinite) number of date variations. %
For instance, there are $70049$ ($365 \times 200 + 49$) dates containing a 2 digit day, a 2 digit month, and a 4 digit year between (inclusive) 1900 and 2100, and if other date (e.g., format, date-part separators, etc.) and visual (e.g., text colour, background colour, typeface, etc.) properties were considered then this number could grow very large.  
For comparison, the large scene-detection MS-COCO benchmark dataset \cite{girshick2014rich} only contains 328k images.
A large number of images would be a problem for traditional computer storage, and AI algorithms that infer continuous parameters may struggle to represent discrete UNIX timestamps. 
We therefore chose not to explore computer vision further in this research.

\begin{figure}[th]

	\includegraphics[width=\linewidth]{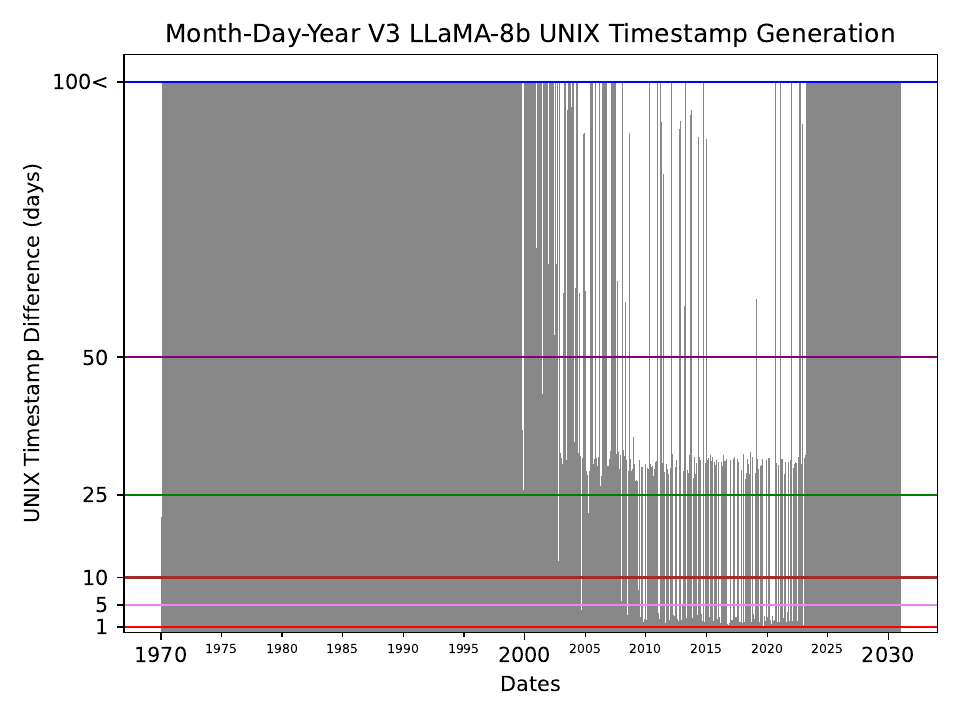}

	\caption{A bar chart that shows the inability of the v3 LLaMA-8b LLM to generate specific UNIX timestamps from explicit prompts.
	The LLM prompt format was ``what is the UNIX timestamp of the date [n]? The date is written in month/day/year format. The UNIX timestamp should be at midnight and in the UTC timezone.''. %
	[n] was a placeholder for each date between the 1st of January, 1970, and the 31st of December, 2031, in the form of a 2 digit (potentially 0-padded) month, a 2 digit (potentially 0-padded) day, and a 4 digit year. 
	The y-axis shows the absolute difference, in days, between the target and predicted UNIX timestamp, and this is capped to 100 days. 
	The x-axis shows dates between the 1st of January, 1970, and the 31st of December, 2030, grouped by year. 
	The generated responses reflect the training data. 
	This suggests that the content generated by an LLM is biased to what it has seen during parameter inference before, and this would be an issue for unique data.
	}
	\label{FIG:direct_llm_date_recognition}

\end{figure}

Our next exploration was of the v3 LLaMA-8b Large Language Model (LLM) produced by Meta \cite{touvron2023llama}, and made accessible by the Ollama\footnote{\url{www.ollama.com}} toolkit.
This illustrates the quint-essential advantages and disadvantages of using a Large-Language Model (LLM) to solve our task \cite{dubey2024llama}, and other LLMs empirically produced similar results.
We configured a generative instance of this LLM with a random seed (`$seed$') of 0,  with the ability for each prompt to generate the same output as the previous one (i.e., $repeat\_penalty$' = 0), and with a temperature (`$temperature$') of 0. %
We provided the prompt 
``what is the UNIX timestamp of the date [n]? The date is written in month/day/year format. The UNIX timestamp should be at midnight and in the UTC timezone.'' 
to the specified v3 LLaMA-8b LLM.
[n] was a placeholder for each date between (inclusive) the 1$^{st}$ of January, 1970 and the 31$^{st}$ of December, 2030.
These contained a 2 digit (potentially 0-padded) month, a 2 digit (potentially 0-padded) day, and a 4 digit year. 
The LLM response format varied (despite attempts to attain a single digit), so we output the absolute of the largest number that it produced. 
Fig.~\ref{FIG:direct_llm_date_recognition} presents these results.
The accuracy of the responses varied significantly depending on the date that was asked, and it demonstrated to us that LLMs were unable to consistently generate precise UNIX timestamps.
This is consistent with other research \cite{liu2024your} too.
The LLM often correctly explained the required methodology (i.e., `a UNIX timestamp is the amount of seconds that have elapsed since the 1$^{st}$ of January, 1970') though, and it may have performed better if it were able to execute external program code.

%% file: sections/capturing_dates_from_machine_written_text.tex
\subsection{The Data}
A total of 1280 dates contained within 884 A4 pages were extracted from 20 different and randomly-chosen medical documents.
These consisted of personal details (name, addresses, etc) and medical histories (ailments, prescriptions, medications, appointments, notes, medications, etc.).
The number of pages in each document varied, and the size and position of the content (i.e., the text-containing-regions) on each page varied.
The pages were orientated either portrait or landscape, the text in each page was orientated either portrait or landscape, and the page backgrounds were typically white. 
It was common for the pages to contain flickers of random noise if the respective document were scanned. 
The text was either machine-written (i.e., mechanically printed) in black ink, or handwritten using blue or black ball-point pen. 
There were, however, cases of handwritten text being written in pencil.
The edges of the text occasionally `bled' (i.e., dilation; \cite{alginahi2010preprocessing}) slightly into the respective page.

The dates were composed of either 2 or 3 parts, and these could be organised horizontally or vertically in a page. %
The dates were read from left to right, and from the top of a page, to its bottom. 
From furtherest to closest: the final date-part was either a 2 or 4 digit year, the penultimate date-part represented a month, and (if present) the part ahead of the month corresponded to a day.
A 2 digit year was in the 21$^{st}$ century if it was less than 40, otherwise it was in the 20$^{th}$ century.
The month could either be a numeric number between (inclusive, and potentially 0-padded to 2 digits) 1 and 12, or a 3 or 4 letter variation of one of the twelve months in the Gregorian calendar: 1) January, 2) February, 3) March, 4) April, 5) May, 6) June, 7) July, 8) August, 9) September, 10) October, 11) November, 12) December.
A day could be between 1 and 28 (inclusive, and potentially 0-padded to 2 digits), and it could be succeeded by an ordinal identifier (i.e., st, nd, rd, th).
Additionally, the number 29 would be a valid day if the month were February and the year were divisible by 4, but not by 400 (i.e., a leap year \cite{newland2024additional}).
29 and 30 would be valid day numbers if the month corresponded to April, June, September, or November.
29, 30, and 31 would be valid if the month corresponded to January, March, May, July, August, October, or December. 
The date ranges were shown using hyphens between two date-parts of the same type (e.g., two months), or words that were analogous to a hyphen (e.g., `[...] to the [...]').
Examples of valid month-year dates are `Jan 2010', `Febr 20' (which would be recognised as February 2020), and `5/2006', examples of valid day-month-year dates are `01/02/2001' and `11th of June, 96', and examples of valid date ranges are `01-02/01/1990' and `March-June 2002'.

\subsection{Data Preprocessing}
\label{ssec:data_preprocessing}
Each page was pre-processed in the following manner before it was transcribed by AI in an attempt to remove noise: 1) images were converted to a single-output-channel greyscale value between 0 (black) and 255 (white), 2) all pixels with a value over 100 were set to white (i.e., binarisation \cite{alginahi2010preprocessing}).

The anticipated number of possible regexes was reduced by removing all instances of `of' from the transcribed text, and the phrases `to the' were replaced by hyphens.

\subsection{Data Annotation}
\label{ssec:data_annotation}
The dates in the documents were manually annotated using the Computer Vision Annotation Tool\footnote{\url{https://github.com/cvat-ai/cvat}} (CVAT) by vetted company employees, and this was performed on company premises, where ISO 27001 and the secure storage of data are rigidly adhered to.

Date annotations without month or year parts were removed from the dataset, and start and end UNIX timestamps were automatically calculated from the annotated date parts.

\subsection{Document Transcription}
\label{ssec:document_transcription}
We began the document transcription by exploring whether the EasyOCR\footnote{\url{https://github.com/JaidedAI/EasyOCR}}, and Tesseract\footnote{\url{https://github.com/tesseract-ocr/tesseract}} \cite{smith2007overview} OCR tools would be able to satisfactorily transcribe the document text, but we found that these were unable to recognise text that was superscripted, rotated, skewed, or handwritten.
We then briefly explored computer vision, but this was not able to recognise the date parts, and this issue was discussed in Sec.~\ref{sec:initial_attempts_to_recognise_dates}.
Next, we explored whether the pdfium\footnote{\url{https://pdfium.googlesource.com/pdfium/}} and poppler\footnote{\url{https://poppler.freedesktop.org/}} tools would be viable, but these struggled to recognise text that was handwritten, or nested within images.
Finally, we deployed a local instance of the open-source LLaVA-13b \cite{liu2024visual}  OCR and HCR tool, and this was able to recognise the majority of document texts using the simple prompt ``what text is in this image?''. %
The recently-developed MegaParse\footnote{\url{https://github.com/QuivrHQ/MegaParse}} library, coupled with Azure's computer vision tools, performed very well when these experiments were repeated.

\subsection{Highly-Accessible Regexes}
\label{ssec:highly_accesssible_regexs}
\begin{figure}[ht]
  \centering
  \includegraphics[width=.6\linewidth]{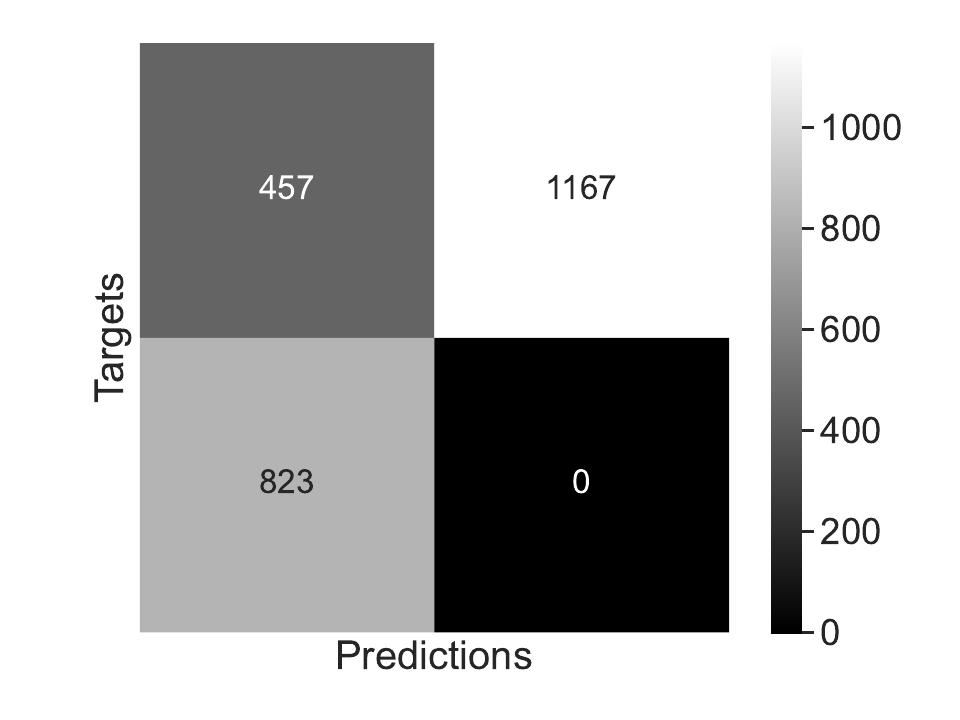}
  
  \caption{The confusion matrix resulting from applying the highly-accessible regexs to the transcribed medical documents. 
  Recall: 28.14\%, Precision: 35.7\%.
  }
  
  \label{fig:community_regexs}
\end{figure}

We began our recognition of the dates and date ranges using 4 easy-to-access  regexes.
This indicated what was possible using the existing tools, despite the inability to extract date parts (and therefore, UNIX timestamps).
The first regex\footnote{\url{https://regex101.com/library/vJ0pI0}} that we used detected patterns consisting of an unrestricted number, a full-length Gregorian calendar month (e.g., January, April, etc.), and a final 4-digit number, each separated by space characters. 
The second regex\footnote{\url{https://regexlib.com/REDetails.aspx?regexp_id=17}} detected patterns consisting of 1 or 2 digits, a further 1 or 2 digits, and a final 4 digits, each seperated by  forward slashes.
The third regex\footnote{\url{https://regexlib.com/REDetails.aspx?regexp_id=235}} detected patterns containing a one or two digit number less than (inclusive) 30, a one or two digit number less than (inclusive) 12, and a two or four digit number, separated by a character from the set [\_-/] (where \_ represents a space). 
The final regex\footnote{\url{(\d{1,2}[-\./](0?[1-9]|1[012])[-\./]((19|20)\d{2}))}} was created by asking the v3 LLaMA-8b chat assistant to ``create a regular expression that can recognise all dates in the day-month-year format''. 
It generated a regex that recognised a one or two digit number, a (potentially 0-padded) number between (inclusive) 1 and 12, and a number between (inclusive) 1900 and 2099, separated by a character from the set \{-.$/$\}.
Applying these regexes to the page transcriptions produced the confusion matrix shown in Fig.~\ref{fig:community_regexs}.
These results indicate that the widely-accessible regexs are not able to achieve our goal.

\subsection{Bespoke Regex}
\label{ssec:besppoke_regexs} 
Next, we used the dates and date ranges contained on 81 pages that were derived from 2 documents to create `optimal' (i.e., those that captured all of the expected text, which we believed would generalise well, which minimised pattern complexities, etc.) date recognition regexes. 
These captured short dates and date ranges that consisted of numerals and seperators from the set [.-\slash{}]  (e.g., xx/xx/xx), and long dates and dates ranges that included other nouns, prepropositions, adjectives, and 3 or 4 letter variations of the Georgian months (e.g., ``the x of x, x''). 
The days of the week (e.g., `Monday', `Tuesday', etc.) were not recognised or validated, dates did not need to explicitly contain a day, and sequential multiline dates (e.g., tables) were assessed. 
Crucially, we were able to identify the individual date parts through a list of keys and values. 
Each regex was mapped to the operations which would need to be performed to the corresponding list of detected dates in order to retrieve each date part. 
For example, the first month part in a day-month-year date was obtained by splitting on the separator (e.g., `/'), and then indexing the middle part of the resultant array.

We treated each recognised date as a range by specifying a start and end UNIX timestamp. 
A day, month, or year part were recognised as a range if the hyphen character (with or without surrounding spaces) were present in addition to another part-separating character.
We considered two date-part ranges together (e.g., the 1st-3rd of May-June, [...]) to be invalid during this research.
The date ranges were implied to go from the first day of the month to the last day of the month if an explicit day were not present, and dates without a hyphen ranged over the 24-hours of a day.

\begin{figure}[ht]
  \centering
  \includegraphics[width=.6\linewidth]{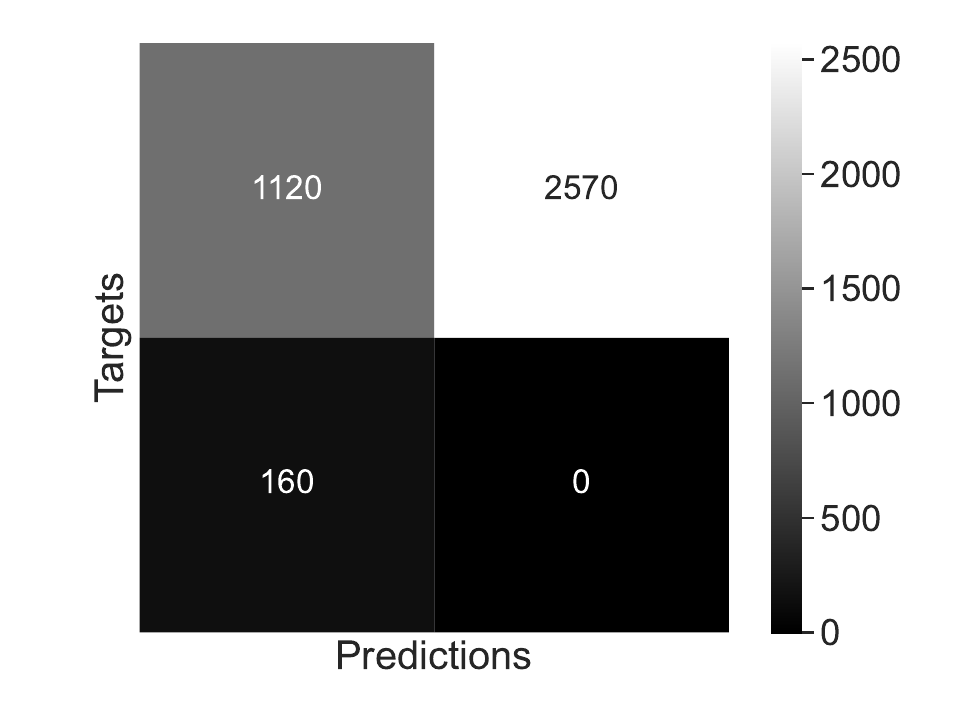}
  
  \caption{The confusion matrix resulting from applying the regexes that were created by us to the transcribed documents. 
  Recall: 30.35\%, Precision: 87.5\%. %
  These were able to reliably capture most of the dates in the transcriptions, but the number of false positives (i.e., detected sequences of text that are not \textit{real} dates) was very large.  %
  }
  \label{fig:bespoke_regexs_confusion_matrix}
\end{figure}

Fig.~\ref{fig:bespoke_regexs_confusion_matrix} shows the confusion matrix resulting from application of the bespoke regex to the transcribed content of the 20 medical documents. 
The bespoke regex was able to detect most (87.5\%) of the real dates and date ranges, but it also detected a lot of dates that did not exist.

\subsection{Regex Synthesis}
\label{ssec:regex_synthesis}
The previous results showed that regexes are a promising way to extract dates and date ranges (and their corresponding parts) from a sequence of text.
However, manual creation of rules (i.e., knowledge engineering \cite{hayes1990construe}) has been known to maximise performance for a long time, and this can be a complex, resource-intensive, error-prone, and unmaintainable process. 
The created regexes would fail to solve the task if the date properties (e.g., language, days per month, months per year, date format, etc.) or requirements (e.g., dayless dates) were to change.

\begin{figure}[ht]
  \centering
  \includegraphics[width=.6\linewidth]{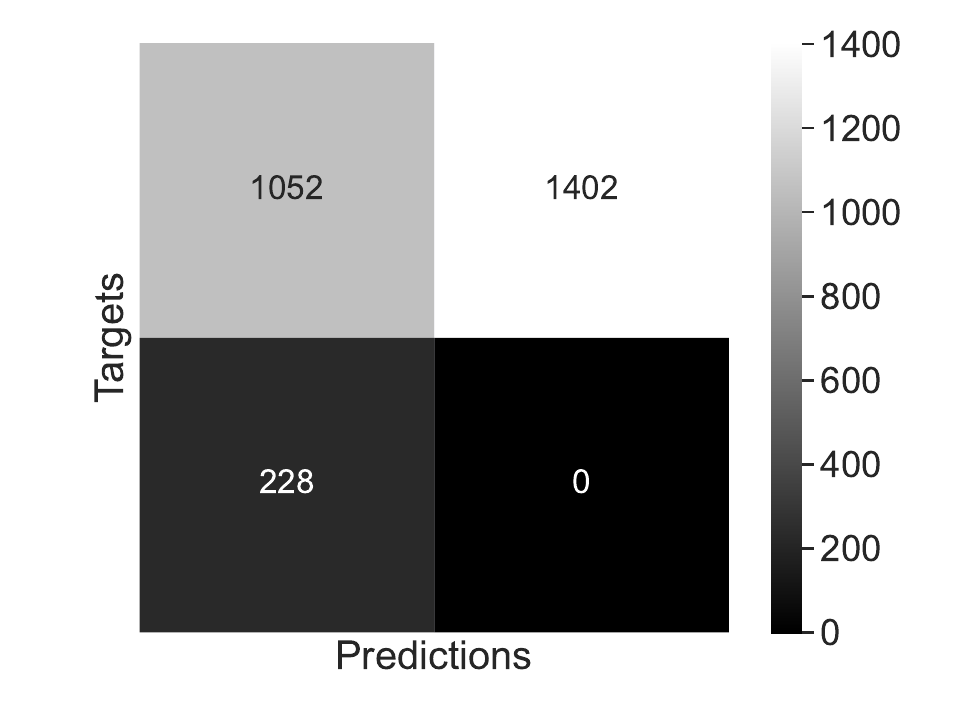}
  
  \caption{The confusion matrix resulting from applying the regexes that were created by reverse-engineering UNIX timestamps to the transcribed documents. 
  Recall: 42.86\%, Precision: 82.96\%. %
  These captured slightly fewer real dates than the bespoke regexes, but they also far fewer dates that did not exist. 
  }
  \label{fig:synthetic_regexs_confusion_matrix}
\end{figure}

Our proposed solution was to automatically create the regexes that would capture all dates between two points in time by reverse-engineering the corresponding UNIX timestamps, and using these to produce sequences of text that can be fed to a regex synthesis algorithm as a collection of positive examples.
The UNIX timestamps have a one-many multiplicity relationship with the date sequences (for example, \textit{03/04/1998} is equivalent to \textit{the 3rd of April, 1998}), which meant that each UNIX timestamp was used to create multiple text sequences, and similarly to before, the regexes were stored in key:value maps that specified how to obtain each date part.
We did not need to generalise (i.e., create negative examples) as we captured the entire input space.
The results of this experiment are shown in Fig.~\ref{fig:synthetic_regexs_confusion_matrix}

The regular expression synthesis algorithm that we used was our implementation of Regex+ \cite{pertseva2022regex}.
The original algorithm creates and stores all possible regexes that would capture a given series of texts in a directed acyclic graph, and the one with the lowest edge cost is chosen.
The graph edges are assigned a variant of the Minimum Description Length (MDL) that balances specificity (i.e., exactness) and simplicity (the number of different patterns that the respective regex could capture).
This was chosen over competing positive-example-only regular expression synthesisers due to the deterministic cost function.

We generated the UNIX timestamps (at midnight) for all dates between the $1^{st}$ of January, 1900, and the $31^{st}$ of December, 2100, extracted the day, month, and year from these, and then used Regex+ to derive regexes from the created text sequences.
These were: 

\begin{itemize}
	\item{a (potentially 0-padded) one or two digit day, a separator, a (potentially 0-padded) one or two digit month, a separator, and a two or four digit year},
	\item{each 3-letter-minimum variation of each long month (e.g., Apr, Apri, April), a long-separator, and a 2 or 4 digit year} 
	\item{a one or two digit day, an appropriate ordinal quantifier (i.e., st, nd, rd, th), the prepositions `of', each 3-letter-minimum permutation of each long month (e.g., Apr, Apri, April), a comma, and a 2 or 4 digit year.} 
\end{itemize}

Sentences containing the day date ranges were constructed by collecting all of the UNIX timestamps in each month, and specifying that each day (potentially 0-padded) ranged to each day greater than itself. 
The months did not have synthesised ranges between them due to the combinatorial explosion that this would have resulted in.
 
The short-separators belonged to the set \{-./\}, and long-separators belonged to the set \{\_-./\}(where \_ represents a whitespace character), and the same separator was used between the month and year, and if it existed, the day and month parts.
We represented date ranges between the day and month digit parts using a hyphen, with and without surrounding spaces, and we removed each detected date and date range after it was detected to stop duplicates.

\subsection{Discussion}
\label{ssec:discussion}
We identified three key research findings:

\begin{itemize}
	\item{That highly-accessible regexes (e.g., those taken from the internet) were not suitable for our date recognition task.} 
	\item{That the regexes which we created for the task of date recognition were able to extract dates and date ranges from text, but these were inflexible to data changes, and they detected many sequences of text that were not actually dates.} 
	\item{That regex synthesis is able to create regexes that are very competitive against those created manually, but these achieved higher recall by sacrificing precision.} 
\end{itemize}

Our research has uncovered three date-recognition considerations which should be explored in the future: 
\begin{itemize}
	\item{contextual dates}
	\item{keyword synonyms (e.g., `and') in regexes},
	\item{date ranges across differing months (e.g., the 3rd of June to the 2nd of July [...]).} %
\end{itemize}
Additionally, these results reveal that regex synthesis should be explored further, as it is a very promising way to build a model that can reliably capture synthetic information.

%% file: sections/acknowledgements.tex
We would like to thank all of colleagues from TMLEP and the University of Kent in Canterbury who were involved in this research.
This research was conducted as part of the ``medical data pagination'' knowledge transfer partnership project (grant number 10048265) that is funded by Innovate UK.

%% file: sections/conclusion.tex
Our research found that it is possible to explainably recognise dates and date ranges in medical documents consisting of text, by transcribing them using modern open-source OCR and HCR tools, and then matching against these using regular expressions.
The existing regular expressions for recognising dates did not meet our consistency and reliability requirements, so we created our own. 
We also needed to create variants that were able to recognise dayless and multi-line dates.
This process was error prone though, as these were able to capture the majority of the real dates, but also a significant amount of text sequences which ``looked'' like dates.
Finally, we showed that it is possible to create regular expressions by reverse-engineering the UNIX timestamps in example dates that we could create.
Our results found that this produced regular expressions that captured fewer dates than in those that were manually created, while capturing far fewer sequences of text which looked like dates.
We therefore recommend further regular expression synthesis exploration in an attempt to further reduce the number of False-positives, to improve the speed at which we are able to automatically create new dates, and to incorporate contextual knowledge.